\documentclass[10pt,twocolumn,letterpaper]{article}
\pdfoutput=1
\usepackage{iccv}
\usepackage{times}
\usepackage{epsfig}
\usepackage{graphicx}
\usepackage{amsmath}
\usepackage{amssymb}
\usepackage{booktabs}
\usepackage{multirow}
\usepackage{arydshln}

\iccvfinalcopy 

\def\iccvPaperID{10341} 

\ificcvfinal\fi

\begin{document}
\title{Seeing through the Mask: Multi-task Generative Mask Decoupling Face Recognition}

\author{Zhaohui Wang\\
Beijing University Of Posts and Telecommunications\\
{\tt\small zhaohuiwang777@gmail.com}
\and
Sufang Zhang\\
Oppo Research Institute\\
{\tt\small zhangsufang@oppo.com}
\and
Jianteng Peng\\
Oppo Research Institute\\
{\tt\small pengjianteng@oppo.com}
\and
Xinyi Wang\\
{\tt\small *@.com}
\and
Yandong Guo\\
{\tt\small Yandong.guo@live.com}
}

\maketitle
\ificcvfinal\thispagestyle{empty}\fi

\begin{abstract}
   The outbreak of COVID-19 pandemic make people wear masks more frequently than ever. 
Current general face recognition system suffers from serious performance degradation, when encountering occluded scenes. 
The potential reason is that face features are corrupted by occlusions on key facial regions.
To tackle this problem, previous works either extract identity-related embeddings on feature level by additional mask prediction, or restore the occluded facial part by generative models.
However, the former lacks visual results for model interpretation, while the latter suffers from artifacts which may affect downstream recognition.
Therefore, this paper proposes a Multi-task gEnerative mask dEcoupling face Recognition (MEER) network to jointly handle these two tasks, which can learn occlusion-irrelevant and identity-related representation while achieving unmasked face synthesis.
We first present a novel mask decoupling module to disentangle mask and identity information, which makes the network obtain purer identity features from visible facial components.  
Then, an unmasked face is restored by a joint-training strategy, which will be further used to refine the recognition network with an id-preserving loss.
Experiments on masked face recognition under realistic and synthetic occlusions benchmarks demonstrate that the MEER can outperform the state-of-the-art methods.
\end{abstract}

\section{Introduction}
\label{sec:intro}
Face recognition (FR) has achieved remarkable progress in recent years \cite{deng2019arcface,hoffer2015deep,2017SphereFace,2018CosFace}, and the accuracy has been continuously improved in most testing datasets \cite{2014Learning,2017VGGFace2,2017AgeDB,2008Labeled,2016Frontal} in general scenarios.
Despite the great performance on normal or slightly occluded faces, state-of-the-art models still struggle under severe occlusions such as masked face recognition (MFR).
\begin{figure}[htp]
\centering
\includegraphics[width=8.3cm]{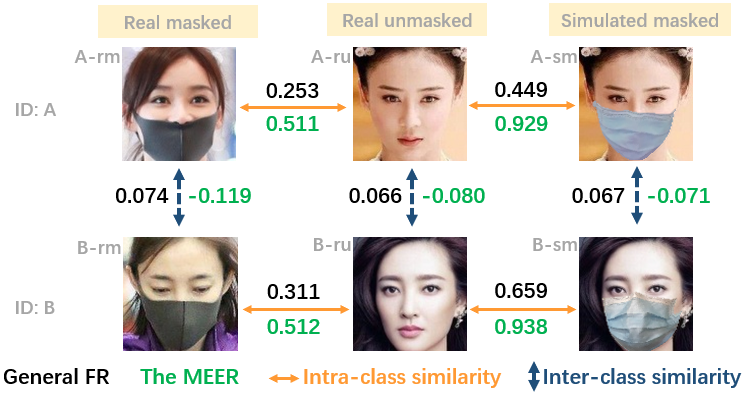}
\caption{\label{fig:mot}Comparisons of similarity between general FR method and the proposed MEER network. The first and second rows are face images from identity A and B. The left and right columns are real and simulated masked face, and the middle is real unmasked face. Masked-unmasked faces similarities within an ID from the MEER (green values) are much higher than those of general FR (black values). And the inter-class similarities of MEER are much lower than those of general FR.}
\end{figure}
Especially today, the COVID-19 pandemic raise the importance and urgency of solving the problem of MFR. 
General FR system cannot extract effective features from mask occluded parts.
The embeddings extracted from masked face will be less discriminative, because masks will increase face intra-class distance and reduce inter-class distance. 
Figure~\ref{fig:mot} gives an example of similarity comparisons between real masked, real unmasked and simulated masked face images from two IDs A and B. 
The values in black are cosine similarities (denoted as cos(.)) between face embeddings from general FR \cite{deng2019arcface}.
It can be seen that, the similarities between real masked and unmasked faces of a same ID (0.253 and 0.311) 
are lower than most of FR system thresholds, which will fail at recognition. 
And the inter-ID similarities of real or simulated masked faces (cos(A-rm,B-rm) and cos(A-sm, B-sm)) are both larger than the similarity between unmasked faces cos(A-ru,B-ru). Therefore, we can conclude that mask occlusions pollute face embeddings and make them less discriminative.

\begin{figure*}[htp]
\centering
\includegraphics[width=16.5cm]{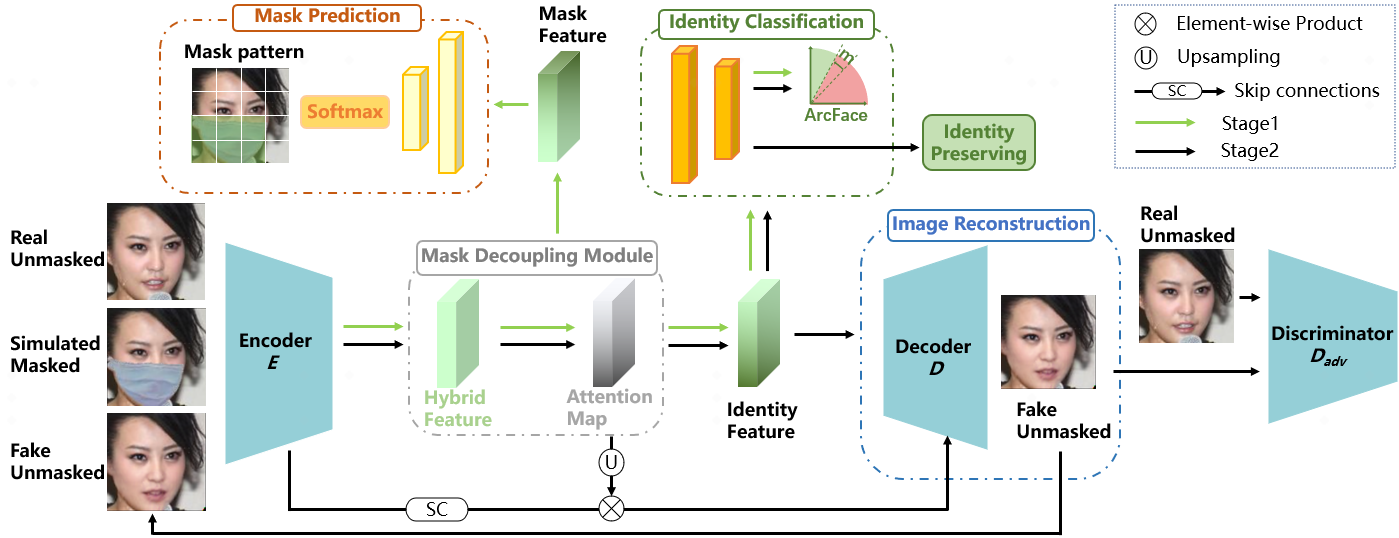}
\caption{\label{fig:pipe}The pipeline of our MEER. 
The green and black arrows indicate the process of multi-task training in stage 1 and joint training in stage 2. 
In stage 1, real unmasked and simulated masked images are input into the encoder $E$ to extract hybrid features.
The hybrid features are weighted by the attention map in the MDM and decoupled into mask and identity features.
The mask and identity features are used to fulfill mask prediction and identity classification tasks respectively. 
In stage 2, both multi-level features of encoder $E$ weighted by the up-sampled attention map and the identity feature will be input to the decoder $D$,
and the fake unmasked images will be reconstructed through the help of discriminator $D_{adv}$.
In addition, the fake unmasked images will be re-fed into the encoder to tune the identity feature by the id-preserving loss.} 
\end{figure*}

Inspired by the fact that human FR system can automatically ignore the occluded part and pay more attention to the rest,
\cite{yuan2022msml,huang2022joint} propose segmentation-based methods, which utilize a network to segment out face occlusion.
However, the mask segmentation results from network appear fuzzy and uncertain, due to the complicated occlusion in the unconstrained scenes.
\cite{qiu2021end2end} model the segmentation problem to a pattern classification, and refine the features polluted by occlusion.
\cite{zhao2022consistent} build consistent sub-decision, which force masked face feature distribution consistent with unmasked face feature as much as possible.

Some GAN-based methods have been proposed to reconstruct the occluded parts \cite{li2017generative,ren2019structureflow,yu2021wavefill,yu2019free,yu2018generative}.
Due to the performance limitation of GAN, the generated image may have ghosts, which will affect recognition accuracy.

To further improve the quality for generative model and provide reliable face identity information,
we propose a multi-task generative mask decoupling face recognition network, termed MEER, to simultaneously learn occlusion-irrelevant identity-related representation and achieve face synthesis.
Specifically, we first decompose the mixed high-level features into two uncorrelated components: identity-related and mask-related features, through an attention module.
We then disentangle these two components in a multi-task learning framework, in which a mask pattern estimation task is to extract mask-related features while a FR task is to extract identity-related feature.
Then the identity-related feature is used to restore a mask removal face image. The new unmasked face will be further fed into the encoder to refine face embedding with an id-preserving loss.
Extensive experiments demonstrate superior performance to existing state-of-the-art methods.
As illustrated in Figure~\ref{fig:mot}, the green values represent face similarities of our MEER. Compared with general FR method (shown in black), the MEER can achieve higher intra-class similarity and inter-class difference, especially in masked face situation.
Our contributions are summarized as follows:

1) We introduce MEER, a multi-task MFR method, which can simultaneously process MFR, mask locating and unmasked face generation tasks.
In MEER, the output of mask locating and unmasked face generation can further help to refine identity information for recognition. 
Experiments on 
masked face test sets demonstrate that our approach accomplishes superior accuracy. 

2) We propose an attention-based 
mask decoupling module to separate the mask-related and identity-related feature from a hybrid high-level feature.
The attention map from the mask decoupling module can further assist the unmasked face generation.

3) We propose a novel joint training strategy with id-preserving loss to achieve delicate face reconstruction and identity preservation, which can further improve the accuracy of MFR.

\section{Related Work}


MFR can be treated as a special scenario of occluded face recognition, which can be divided into three mainstream types. 
The first is to refine face embeddings by adding more masked face images, or obtaining attention maps which are robust to occlusions.
The second is to purify features polluted by masks with mask segmentation as training supervision.
The third focuses on repairing the occluded area on the face. 
We will give a brief introduction about these methods in the following subsections.

\subsection{Feature Refinement Occlusion FR}
\cite{wang2021mask} simply generated masks on unmasked faces by face landmarks, and chose several backbones to extract embeddings with different FR losses.
\cite{feng2021towards} generated simulated masked face and trained a FR model together with unmasked real faces, which achieved the superior performance in a MFR competition.
\cite{zhao2022consistent} first extracted face feature, which was further input to consistent sub-decision network.
Then a bidirectional KL divergence constraint was applied to constrain unmasked feature and sub-decisions feature for optimizing the sub-decision consistency.
\cite{wang2021dsa}, \cite{wang2021aan} and \cite{wang2022cqa} designed different approaches to generate diverse attention maps of a face image. 
Then all attention maps were employed to refine local and global face features, and these features were merged to a face embedding, which is robust to occlusions.
\cite{yin2019towards} proposed spatial activation diversity loss and feature activation diversity loss to learn structured feature response which was insensitive to local occlusions.
\cite{ding2020masked} proposed that partial and global branch to learn discriminative partial and global feature.
\cite{zhang2022learning} used an upper patch attention module to extract the local features and adopt dual-branch training strategy to capture global and partial feature to achieve MFR.
\cite{huber2021mask} built a knowledge distillation model for MFR, where a teacher model learns FR and the student model learns MFR by simulated masked face data augmentation with teacher's supervision.


\subsection{Segmentation Based Occlusion FR}
\cite{qiu2021end2end} proposed a feature pyramid extractor
to fuse multi-scale features and a fine-grained occlusion pattern predictor to remove the polluted feature. 
However, this method cannot accurately express various real world occlusion, resulting in degenerating recognition performance.
To solve the above problem, \cite{huang2022joint,yuan2022msml} proposed a feature refinement method based on segmentation. 
In the \cite{huang2022joint}, the extracted features were divided into two channels, one for prediction segmentation label and the other for masked face feature purification. 
Specially, the segmentation label was feed into a channel refinement network to get a occlusion mask feature with mask position information. 
They decoupled the occlusion mask feature from original face feature by using a feature purification module.
Similar with \cite{huang2022joint}, \cite{yuan2022msml}
also used a occlusion segmentation branch to predict a segmentation map. 
They utilized multiple feature masking modules to achieve purified multi-scale masked features.
Each feature masking operator received facial features and occlusion segmentation representations at the corresponding layer, and outputted purified facial features.
\cite{song2019occlusion} proposed a pairwise differential siamese network named PDSN, which divided the face into patches as \cite{qiu2021end2end}. The mask pattern was predicted by several PDSN networks.
Then a mask dictionary was established accordingly, which was used to composite the feature discarding mask for removing the polluted features.

Segmentation based method usually needed an extra segmentation network, which increased the network computation and model complexity.
Sub-decision based and segmentation based  methods both adapted a masked face feature purify module to remove the polluted feature by mask, which may abandoned some features in visible areas and resulted in indelicate removing.

\subsection{Inpainting Based Occlusion FR}
The early masked face inpainting method \cite{zhao2017robust} adopted an encoder LSTM and a decoder LSTM to generate unmasked face.
Benefiting from the remarkable progress in GAN (Generative Adversarial Networks), \cite{li2017generative} utilized a global GAN to achieve coarse face generation and a face parsing network with local discriminator to achieve fine grained face generation.
\cite{cai2020semi} used a occlusion aware module to predict occlusion mask.
Then a face completion module was employed to restore face image.
\cite{din2020novel} combined segmentation module and GAN to achieve mask removal.

Some face inpainting methods reconstructed the missing facial components by using the context information around the occluded region.
\cite{yu2018generative} proposed a coarse network with a refinement network to reconstruct the masked parts. The contextual attention layer learned to borrow feature information from known background patches to generate missing parts.
\cite{yu2019free} proposed a gated convolution and SNPatchGAN loss to restore the unmasked face.
\cite{yu2021wavefill} translated images from spatial domain to three different frequency domains to achieve face inpainting.

\section{Method}

The framework of the  multi-task generative mask decoupling face recognition (MEER) network is illustrated in Figure~\ref{fig:pipe}.
The proposed method contains two training stages: (1) face feature extraction and disentanglement (green arrows in Figure~\ref{fig:pipe}); (2) joint training of face generation with mask removal and feature refinement (black arrows in Figure~\ref{fig:pipe}).
These two stages will be elaborated in \ref{stage1} and \ref{stage2} respectively.

\begin{figure}[htp]
\centering
\includegraphics[width=6cm]{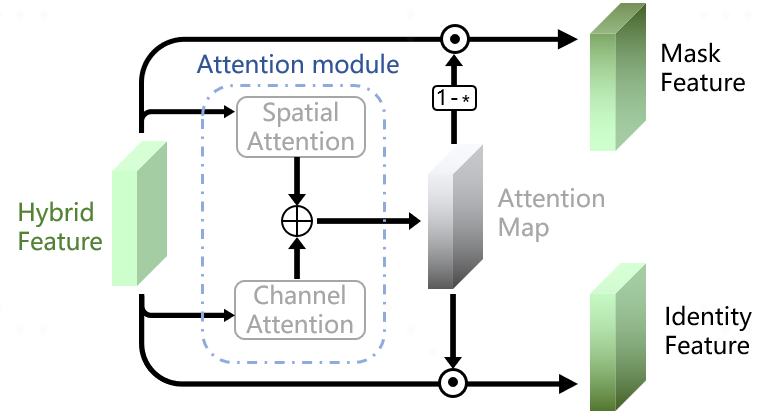}
\caption{\label{fig:MDM}The MDM of our MEER.}
\end{figure}

\subsection{Stage 1: face feature extraction and disentanglement}
\label{stage1}
In this stage, face images $I$ will be input into a encoder $E$ to extract their features. And $I$ contains two face image sets: real unmasked images $I_{\textmd{ru}}$, and simulated masked images $I_{\textmd{sm}}$ generated by FaceX-Zoo \cite{feng2018joint}. 
The output features contain face identity information and other id-unrelated information, thus we name it hybrid features $X=E(I)$.
Then, we build a mask decoupling module (MDM) to disentangle face identity feature $X_{{id}}$ and mask feature $X_{{mask}}$ from $X$. The detail of MDM will be introduced later.
Finally, these two branches are adopted to jointly learn FR and mask location prediction.

For FR task, $X_{id}$ is fed into several linear layers $M^1$ and converted into identity embedding $Z_{{id}}= M^1(X_{{id}})$. Finally, ArcFace\cite{deng2019arcface} is employed as the identity classification loss.
For mask position prediction task, $X_{mask}$ is fed into several linear layers $M^2$ and converted into mask embedding $Z_{mask}= M^2(X_{{mask}})$. 
We adopt softmax classifier proposed by \cite{qiu2021end2end} to predict mask locations. Specifically, face image is divided into a grid map. Different patch combinations containing mask represent different mask patterns, where 101 types of patterns are used in this loss.
The mask positions pre-marked in map are used as the classification ground truth, which is also generated by FaceX-Zoo without manually annotation.
The loss function in stage 1 can be formulated as follows:
\begin{equation}
\label{l_multi_task}
    \mathcal{L}_{MEER-s1} =  l_{sm}(Z_{mask}, y_{mask}) + \lambda l_{Arc}(Z_{id}, y_{id})
\end{equation}
where $y_{mask}$, $y_{id}$ denote the ground truth of mask position pattern and identity label, $\lambda$ is a hyper-parameter for balancing two tasks.

\subsubsection{Mask decoupling module}

Feature decoupling is a widely used approach in face tasks under specific scenarios, such as age-invariant FR, and face anti-spoofing. Inspired by \cite{huang2021age}, we build the mask decoupling module (MDM), which decouples the hybrid feature in the high-level semantic space through the residual attention, as shown in Figure~\ref{fig:MDM}.
First, spatial \cite{woo2018cbam} and channel \cite{hu2018squeeze} attention will be generated from hybrid feature $X$ by several light-weighted convolution layers.
Then the spatial and channel attention will be merged into an mask attention map.
At last, the identity feature and mask feature will be separated by this attention map.
%
%
\begin{figure}[htp]
\centering
\includegraphics[width=7cm]{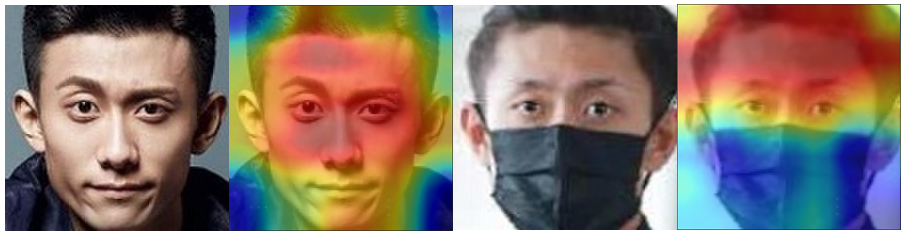}
\caption{\label{fig:attmap}The example of the attention maps on unmasked and masked faces. Note that red and blue areas represent high and low attentions.}
\end{figure}
The formula of feature decoupling is as follows:
\begin{equation}
    X_{{id}} = X \odot \Phi(X), \thinspace X_{{mask}} = X \odot (1 - \Phi(X))
\end{equation}
\begin{equation}
    X = X_{{id}} + X_{{mask}}
\end{equation}
where $\odot$ represents Hadamard product; $\Phi$ represents the attention module, and $\Phi(X)$ is the output attention map.
An example of the attention maps on both masked and unmasked image is shown in Figure~\ref{fig:attmap}.
It can be seen that, when the input face is unmasked, the attention is on the whole face. 
However, when the input is a face with mask, the attention focuses on the uncovered facial parts, such as eyes and brows.
As a result, for an unmasked face, its identity feature includes information of the whole face, and the non-face area (such as shoulders) will be suppressed. But for a masked face (no matter it is simulated or real), its identity feature only focuses on the facial part without mask occlusion. 

Note that the MDM can bring purer identity information to FR branch, which will achieve a higher recognition performance.
In addition, pure identity feature will pave the way for the task of face generation with mask removal in the next section.


\subsection{Stage 2: MEER joint training}
\label{stage2}
In this stage, we manage to do two things: 1, restoring an unmasked face if the input is a masked face; 2, adopting generated face to refine the identity feature for greater performance of MFR. 

First, for a masked face $I_{\textmd{sm}}$, we try to restore an unmasked face while keeping its identity. 
As shown in Figure~\ref{fig:pipe}, the identity feature $X_{id}$ from masked face is fed into a decoder $D$ for image reconstruction.
The output will be a fake unmasked face, denoted as $I_{\textmd{fu}}$.
For the unmasked face $I_{\textmd{ru}}$, its feature $X_{id}$ will not be fed into the decoder and no face needs to be restored.
In order to generate face with more details, we build multi-level connections from $E$ to $D$, similar with the structure of U-net.
This skip connections (SC) will deliver multi-scale features extracted by the encoder to the decoder, which leads to a better generation.
\begin{figure}[htp]
\centering
\includegraphics[width=8.3cm]{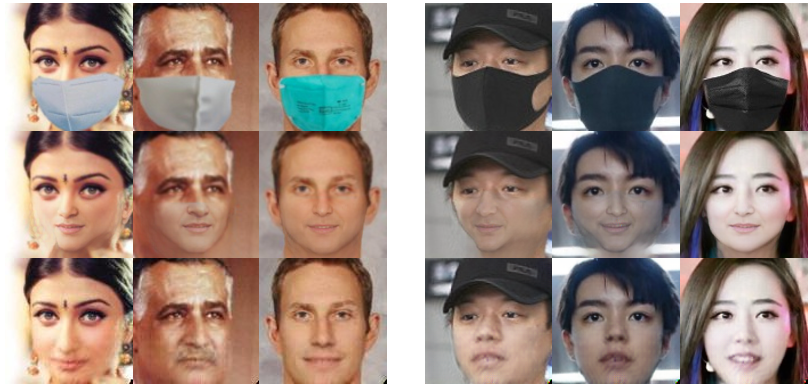}
\caption{\label{fig:visesult} Mask removal results comparisons. The left and right parts are results from simulated masked faces of MLFW and real masked faces of RMFD. The first line is the masked images, 
the 2-nd and last lines are the inpainting results from \cite{yu2021wavefill}  and our MEER generated results.}
\end{figure}
To make sure the information in SC do not contain mask feature and avoid mask artifacts appearing on the generated face, we design a mask information suppression process to purify the multi-level connection information by merging the attention map $\Phi(X)$ in the MDM to the SC.
Formally, the process can be written as:
\begin{equation}
    f_{{l}}^{'} = f_{{l}} \thinspace \odot \thinspace U_{{l}}(\Phi(X))
\end{equation}
\begin{equation}
    I_{\textmd{fu}} = D( \{ f_{{l}}^{'} \}_{{l=1}}^{3}, X_{{id}})
\end{equation}
where $f_{{l}} \lvert l={1,2,3}$ denotes features extracted from the $l$-th layer of the encoder $E$.
$U_{{l}}(\cdot)$ represents process of up-sampling to the size of the $l$-th feature.
$f_{{l}}^{'}$ is the $l$-th level of SC output, which will be merged into decoder.

For further improving the quality of restored face, we use a discriminator $D_{{adv}}$ from the GAN training strategy together with the decoder $D$. 
Here, we adopt the Patch Discriminator.
In order to make training easier, we apply pixel loss as a constraint between $I_{\textmd{ru}}$ and $I_{\textmd{fu}}$.
The GAN training process in stage 2 can be formulated as follows:
\begin{equation}
    \mathcal{L}_{{D}} = \frac{1}{2} \mathbb{E}_{I_{\textmd{fu}}}[D_{{adv}}(I_{\textmd{fu}} )-1]^{2}
\end{equation}
\begin{equation}
\mathcal{L}_{D_{adv}} = 
\frac{1}{2} \mathbb{E}_{I_{\textmd{ru}}}[D_{{adv}}(I_{\textmd{ru}})-1]^{2} 
+ \frac{1}{2} \mathbb{E}_{I_{\textmd{fu}}}[D_{{adv}}(I_{\textmd{fu}})]^{2}
\end{equation}
\begin{equation}
    \mathcal{L}_{{rec}} = \Vert I_{\textmd{fu}} - I_{\textmd{ru}} \Vert_2^{2}
\end{equation}

\begin{table*}
\setlength\tabcolsep{1pt}
\begin{center}
\begin{tabular*}{0.9\linewidth}{ccccccccc}
\toprule
\multirow{2}{*}{Method} & \multicolumn{2}{c}{\multirow{2}{*}{MLFW}} & \multicolumn{2}{c}{\multirow{2}{*}{RMFD}} & \multirow{2}{*}{MFR2} & \multicolumn{3}{c}{LFW-masked} \\ 
& & & & & & Face-Face &Face-Mask &Mask-Mask \\
\cmidrule(lr){2-3} \cmidrule(lr){4-5} \cmidrule(lr){7-9}
& (ACC\%) & (AUC\%) & (ACC\%) & (TPR@FAR=1\%) & (ACC \%) &  \multicolumn{3}{c}{(AUC\%)} \\
\midrule
CQA-Face\cite{wang2022cqa} & 92.78 & - & - & 34.22 & - & - & - & - \\
DSA-Face\cite{wang2021dsa} & 92.91 & - & - & 39.21 & - & - & - & - \\
AAN-Face\cite{wang2021aan} & 92.91 & - & - & 37.18 & - & - & - & - \\
INFR\cite{yin2019towards}&-&-&-&-&93.52&99.69&97.92&97.95\\
LPD\cite{ding2020masked}&-&-&-&-&92.60&98.76&98.38&98.15\\
UPA\cite{zhang2022learning}&-&-&-&-&95.22&99.58&99.41&99.37\\
MaskInv \cite{huber2021mask} & - & 93.65 & - & \textbf{84.50} & 99.62 & - & - & -\\
CS \cite{zhao2022consistent}              & - & 93.40 & -& 79.11 & \textbf{99.75} & 99.75 & 99.81 & 99.18  \\
FROM\cite{qiu2021end2end}  & 92.55* & 90.41* & 91.36* & 30.67* & 96.22 & \textbf{99.93*} & 99.52* & 99.52* \\
JS\cite{huang2022joint}  & 90.77* & 94.80* & 80.74 & 61.05* &	95.40* & \textbf{99.92*} & 99.76* & 99.67* \\
MSML\cite{yuan2022msml}   & 93.45* & \textbf{96.74*} & 92.91*  & 83.09* & 95.98* & 99.88* &  \textbf{99.85*} &  \textbf{99.81*} \\
\hline
ArcFace-MaskAug & 93.69 & 96.42 & 92.82 & 83.84 & 99.64 &99.86 &99.81 & \textbf{99.80} \\
Without MDM      &93.02 &96.25 & 92.49   & 81.78 &99.41 & 99.87 &99.79&99.75 \\
MEER-stage1      &  \textbf{94.05} & 95.57 & \textbf{93.09} & 84.32 & 99.52 & 99.87 & \textbf{99.83} & 99.77 \\
MEER-stage2      & \textbf{93.70} & \textbf{96.59} &  \textbf{93.61}  & \textbf{86.55}  & \textbf{99.76}  & \textbf{99.92} &  \textbf{99.83} & 99.77  \\
\bottomrule
\end{tabular*}
\caption{ 1:1 verification on MLFW, LFW-masked, RMFD and  MFR2 datasets. * marks the results reproduced by us. The bold-text values show the top two results. - represents that no results are provided in original papers and no official open-source is released.}
\label{tab_fwsc}
\end{center} 
\end{table*}

Generating unmasked face is a straight-forward idea to solve the problem of MFR, such as \cite{din2020novel}.
However, these generative methods are based on inpainting, and thus excessively rely on the accuracy of mask segmentation. These methods can not preserve the ID of restored face from the original face. 
To ensure the identity consistency between generated fake unmask image $I_{\textmd{fu}}$ and original real unmask image $I_{\textmd{ru}}$, $I_{\textmd{fu}}$ will be re-input into the encoder $E$ to form a joint training structure, as shown in Figure~\ref{fig:pipe}. 
An id-preserving loss $\mathcal{L}_{{id-preserving}}$ is designed to force the identity of fake unmasked face $Z_{{id}}^{I_{\textmd{fu}}}$ similar with its corresponding real unmasked face $Z_{{id}}^{I_{\textmd{ru}}}$, which is shown as follows:
\begin{equation}
\label{l_idp}
    \mathcal{L}_{{id-preserving}} = \textmd{sim}(Z_{{id}}^{I_{\textmd{fu}}}, \thinspace Z_{{id}}^{I_{\textmd{ru}}})
\end{equation}
where $\textmd{sim}(.)$ donates the cosine distance between two embeddings.
In stage 2, ArcFace is also used to supervise the identity of all images from $I_{\textmd{ru}}$, $I_{\textmd{sm}}$ and $I_{\textmd{fu}}$. 
Thus an addition identity loss will be adopted on $I_{\textmd{fu}}$ as follows:
\begin{equation}
    \mathcal{L}_{id}^{'} = l_{Arc}(Z_{id}^{I_{\textmd{fu}}}, \thinspace y_{id})
\end{equation}
where $y_{id}$ of a fake unmasked face should be identical with its corresponding real unmasked face. Finally, the full loss of our MEER in stage 2 can be written as follows:
\begin{equation}
\label{l_total}
\begin{aligned}
    \mathcal{L}_{MEER-s2} = & \thinspace \mathcal{L}_{D} + \alpha \mathcal{L}_{adv} 
    +  \beta (\mathcal{L}_{id}  + \mathcal{L}_{id}^{'}) \\
    &  + \gamma \mathcal{L}_{rec} + \eta \mathcal{L}_{id-preserving}
\end{aligned}
\end{equation}
where $\alpha$, $\beta$, $\gamma$ and $\eta$ are the hyper-parameters.

\section{Experiments}
\textbf{Training Datasets.}  
MS-Celeb-1M \cite{guo2016ms} is a mainstream dataset used in large-scale FR, which consists of 100K identities (each identity has about 100 facial images).
MS1M-v2 is a clearer subset of MS-Celeb-1M that contains 5.8M images from 86K classes.
We build a new dataset: MS1M-v2-Aug as the training dataset for MFR.
In specific, we use FaceX-Zoo \cite{wang2021facex} to generate simulated masked face images from 25\% of images in the MS1M-v2. 
MS1M-v2-Aug consists of 86K ids and approximate 7M face images with both real unmasked and simulated masked faces.

\textbf{Testing Datasets.}  To comprehensively evaluate the performance of different MFR methods, 
we categorize the testing datasets into two types: 
real masked face datasets (RMFD, MFR2) and simulated occlusion face datasets (MLFW, LFW-masked).
RMFD \cite{huang2008labeled} contains 4015 raw images of 426 persons and constructs about 3000 pairs of faces from same identities and different identities. 
MFR2 \cite{anwar2020masked} contains 848 positive and negative pairs from 53 identities of celebrities and politicians. The dataset has 269 images that are collected from the internet.
MLFW \cite{wang2021mlfw} contains 6000 face pairs with one real unmasked face and one simulated masked face.
LFW-masked \cite{zhang2022learning} is a simulated-masked face dataset based on LFW \cite{huang2008labeled}. 
This dataset is separated in three different scenarios \cite{zhao2022consistent}, which are face to face, face to mask and mask to mask, and every scenario contains 6000 verification pairs.


\textbf{Evaluation Metrics.} We use 1:1 face verification accuracy (ACC), TPR@FAR(1\%) and AUC as metrics to evaluate the MEER and the state-of-the-art methods.

\subsection{Implementation Details}
\textbf{Prepossessing.} 
We choose five landmarks (two eyes, nose and two mouth corners) to achieve aligned face with size 112$\times$112.   
Following \cite{qiu2021end2end,huang2022joint}, we normalize pixel values to [-1.0, 1.0] in training and testing.

\textbf{Network Structure.} 
For fair comparison, we adopt IResNet-50 as the encoder $E$, following the backbone choice of \cite{qiu2021end2end,zhao2022consistent,huang2022joint}.
The structure of decoder $D$ is a GAN similar with pix2pix, which has a style encoder and then inserts the style into some of the convolutional layers through adain. 
The discriminator $D_{adv}$ is a patch discriminator \cite{isola2017image} which penalizes the framework for better visual quality.


\textbf{Training.} The training of MEER contains two stages.
Specifically, in stage 1, 
simulated masked face $I_{sm}$ and real unmasked face $I_{ru}$ are randomly fed into network.
Both $I_{sm}$ and $I_{ru}$ conduct disentanglement by MDM.
Stage 2 is a joint training process of $E$, $D$ and $D_{adv}$. 
Different from stage 1, we send paired masked and unmasked data to the network, and the generated unmasked face also need to be sent to the encoder.
%
%
The network is trained on 8 Tesla V100 GPUs with a batch size
of 128 and a momentum of 0.9.
We employ 
Adam as the optimizer. The weight decay is set to $5e^{-4}$.
The learning rate begins with 0.01, and it is divided by 10 
until 0.0001. 
The hyper-parameters $\lambda$, $\alpha$, $\beta$, $\gamma$ and $\eta$ are set to 0.01, 1, 1, 10 and 0.1, respectively.


\subsection{Evaluation on mask datasets}
As the test protocols of previous methods are inconsistent, we reproduce the experimental results of several state-of-the-art MFR works \cite{qiu2021end2end,yuan2022msml,huang2022joint} for fair comparison.
In Table\ref{tab_fwsc}, metrics with * represent they are reproduced by authors' release code. 
\textbf{MEER-stage1} means the model trained as section \ref{stage1}.
\textbf{MEER-stage2} means the final model obtained by the joint-training procedure on stage 2, following the training of stage 1.
Note that the results of MEER-stage2 in Table\ref{tab_fwsc} is evaluated only by inference the encoder (without face generation or re-feeding restored images). 
We evaluate ACC on MLFW, RMFD, MFR2; TPR on RMFD and AUC on MLFW and LFW-masked.
In Table\ref{tab_fwsc}, the top two results are thickened.
Generally, our MEER outperforms the prior works and achieves the top two results in most of datasets.
First, we can conclude that MEER-stage2 can surpasses MEER-stage1 on almost all test sets. 
The excellent performance of MEER-stage2 benefits from the joint training strategy and reconstructed unmasked image which is re-fed into encoder. Furthermore, the id-preserving loss improves the recognition accuracy.

Comparing with the mainstream MFR methods in Table\ref{tab_fwsc},  
our method shows significant advantages.
MEER adopts mask pattern prediction and multi-task training strategy similar with FROM \cite{qiu2021end2end}.
However, MEER adopts MDM, which achieves a better disentangling of identity and mask features.
In addition, MEER uses the reconstructed images to tune the encoder, which further improves the performance of MFR.
Except the Face-Face scene of LFW-masked, our evaluation results on all test datasets significantly surpasses the FROM.
JS \cite{huang2022joint} and MSML \cite{yuan2022msml} use segmentation network for mask prediction.
They adopt single or multiple disentangling modules to purify identity feature.
\begin{figure}[htp]
\centering
\includegraphics[width=8cm]{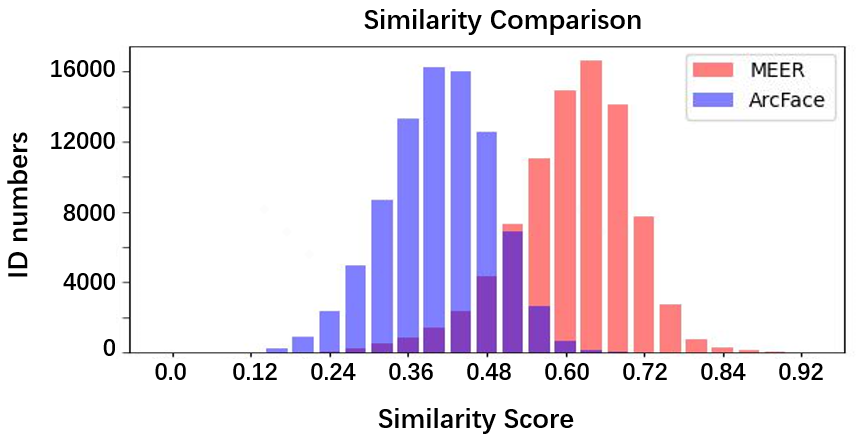}
\caption{\label{fig:sim}Intra-class masked-unmasked face similarity distribution on MS1M-v2-Aug of MEER and ArcFace. The horizontal and vertical ordinates of this chart represent the cosine similarities of a ID, and the number of IDs in each interval.}
\end{figure}
In some complicated scenarios, such as large poses or severe occlusions, fuzzy or incomplete segmentation maps may appear.
It is risky to use noisy segmentation labels to purify facial feature, which will result in recognition performance degradation.
MSML, JS and MEER are all trained on the simulated masked face images.
The segmentation labels of MSML are more accurate on the simulated datasets than real mask datasets, so MEER and MSML obtain competitive results in the simulated datasets (e.g. MLFW and LFW-masked).
Our MEER performs better in real masked face scenarios (e.g. RMFD and MFR2).

The consistent sub-decision network CS \cite{zhao2022consistent} adopts dropout blocks to extract subspaces, which directly remove some masked area features.
CQA-Face \cite{wang2022cqa}, AAN-Face \cite{wang2021aan} and DSA-Face \cite{wang2021dsa} design several branches to get different local facial features for general occlusion situations.
 INFR \cite{yin2019towards}, LPD \cite{ding2020masked}, UPA\cite{zhang2022learning} focus on partial structured feature, which depends on the learning ability of the local branch.
Those methods focus on looking for different local distinguishing features, which confine to the numbers of local branches or the precision of local features.
However, we can get more pure global feature to reach a higher performance through MDM.
%
%
MaskInv \cite{huber2021mask} built a teacher and student module, where MFR is only trained on student network and heavily depend on masked data argumentation.
Our method can obtain more accurate attention map in mask scenarios, and thus it surpasses MaskInv on most of metrics.


Benefiting from the mask-decoupled pure facial features, MEER can generate realistic unmasked face. 
We exhibit the restored unmasked images from \cite{yu2021wavefill} and ours in the Figure~\ref{fig:visesult}.
The first line gives simulated masked faces of MLFW and real masked
faces of RMFD.
The second line is \cite{yu2021wavefill}, which can repair the complete face. However, the repaired occluded area has more artifacts than our method, and the restored area is almost smiling face which lacks of diversity.
Different from \cite{yu2021wavefill}, the MEER does not need mask segmentation map as restoration supervision. 
We employ mask information suppression with both adversarial and reconstruction loss in the process of image reconstruction. Therefore our mask removal results contain less artifacts.

In the training of the MEER, we don't explicitly constrain the similarity of masked and unmasked faces (note that the id-preserving loss in Eqn.\ref{l_idp} only focuses on real and fake unmasked faces).
However, as shown in the distribution of similarity on the training set in Figure~\ref{fig:sim}, the proposed MEER increases the intra-classes similarities between masked and unmasked faces.

\begin{table}[htp]
  \begin{center}
    \begin{tabular}{ccccc}
    \toprule
    \textbf{Method} & \textbf{RMFD} & \textbf{LFW} \\
    \midrule
    No SC & 92.35 & 99.70 \\
    One SC & 92.84 & 99.70 \\
    Three SC & 93.02 & 99.70 \\
    Three SC + MIS & \textbf{93.60} & \textbf{99.73} \\
    \bottomrule
    \end{tabular}
    \caption{The accuracy(\%) of different settings of skip connections (SC) and mask information suppression (MIS)}.
    \label{tab:table4}
  \end{center}
\end{table}

\subsection{Evaluation on general FR datasets}

To verify the robustness of the proposed method in general FR datasets,
we perform a 1:1 face verification as shown in Table\ref{tab:table3}.
As shown in this table, comparing with FROM, JS, our method shows better generalization in unmasked FR.
CQA-Face, DSA-Face, and AAN-Face obtain higher results, since they only used unmasked face images while training. 
MSML and MarkInv achieve similar performance with ours on different datasets.

\begin{table}[htp]
\setlength\tabcolsep{3pt}
  \begin{center}
    \begin{tabular}{ccccc}
    \toprule
    \textbf{Method} & \textbf{LFW} & \textbf{AgeDB} & \textbf{CFP-FP} & \textbf{IJB-C} \\
    \midrule
    CQA-Face   & 99.83 & - &	98.49 &	- \\
    DSA-Face   & 99.85 & - &	98.69 &	95.51 \\
    AAN-Face   & 99.87 & 98.15 &	98.63 &	- \\
    FROM & 99.38 & - &	- & -	 \\ 
    JS & 99.48 & 94.65 & 96.10 & 80.82 \\
    MSML & 99.83 & 97.28 & 96.17 & 93.94 \\
    MaskInv & 99.82 & 97.83 & 97.53 & - \\
    
\hline
    MEER-stage1 & 99.75 & 97.37 & 96.57 & 94.38 \\
    MEER-stage2 & 99.73 & 97.15	& 96.63 & 94.90 \\
    \bottomrule
    \end{tabular}
    \caption{1:1 verification results on LFW, AgeDB, CFP-FP and IJB-C datasets of different MFR methods. 
    }
    \label{tab:table3}
  \end{center}
\end{table}

\subsection{Ablation Studies}
We perform ablation experiments on different combinations of key modules and parameters applied in the paper. 

\begin{figure}[htp]
    \centering
    \includegraphics[width=7.5cm]{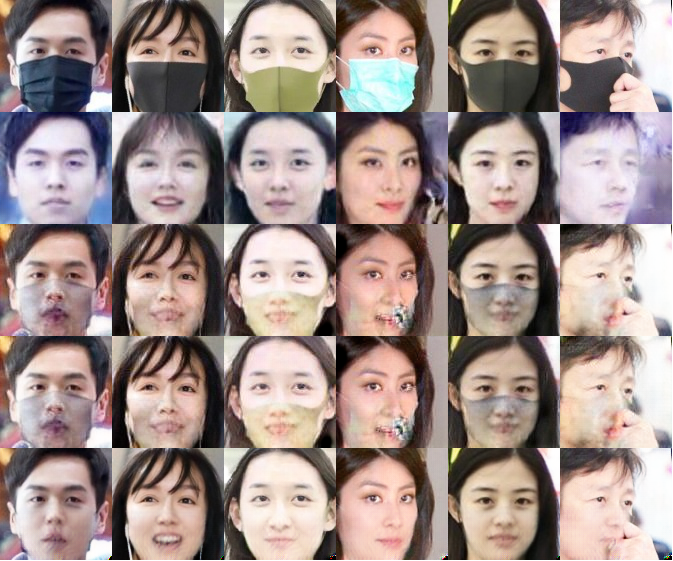}
    \caption{The examples of face mask removal applying different combinations of SC and MIS. From the top to the bottom, they are the original masked images, the generative unmasked face images of no SC, one SC, three SC and three SC with MIS.}
    \label{fig:AS_rmfd}
\end{figure}

First of all, we shows the effectiveness of MDM in Figure~\ref{fig:MDM}.
ArcFace-MaskAug in Table\ref{tab_fwsc} represents an IResNet-50 model trained by ArcFace loss on MS1M-v2-Aug, similar with \cite{feng2021towards}. 
The difference between ArcFace-MaskAug and MEER-stage1 is that we additionally adopt MDM  and multi-task training with mask prediction in Eqn.\ref{l_multi_task}.
Comparing the MEER-stage1 and ArcFace-MaskAug, we can find that the MDM significantly improves the accuracy by 0.36\% and 0.27\% on MLFW and RMFD respectively, and achieves competitive results in other benchmarks.
\textbf{Without MDM} represent general feature disentanglement method to learn mask location and FR, where half channel feature of X is used for FR and the other half is for mask pattern classification.
Comparison between MEER-stage1 and Without MDM also proves the effectiveness of MDM.

Then, we verify the importance of the skip connections (SC) between encoder and decoder.
Illustrated in Table\ref{tab:table4}, we provide the 1:1 verification accuracy comparison of different SC settings on RMFD and LFW.
As the number of SC used in image reconstruction step becomes larger, the accuracy on RMFD also gets larger (from 1-st to 3-rd rows).
When attention mask is used on SC to suppress the mask information (the 4-th row), its accuracy is 0.58\% and 0.03\% higher than the settings of three SC on RMFD and LFW. 

We also gives some mask removal results with different SC settings in Figure~\ref{fig:AS_rmfd}. 
The 1-st line is real masked images. 
The 2-nd line is generative unmasked faces without SC.
Because the generated image uses the identity feature purified by MDM, the image is not polluted by mask information. 
However, due to the multi-layer encoding of $E$, the details of input images are seriously lost, 
such as skin color, hairstyle and background. 
The 3-rd and 4-th lines are reconstruction results with one and three SC.
By adding SC, more image information is retained in no occluded area. 
But there are some mask artifacts around mouth and nose areas, because the mask information is included when SC are used.
As shown at the bottom line, this problem has been greatly alleviated after mask information suppression is used.

\begin{figure}[htbp]
    \centering
	\begin{minipage}{0.45\linewidth}
		\centering
        \begin{tabular}{cccc}
        \toprule
        \textbf{$\gamma$}  & \textbf{RMFD} & \textbf{LFW} \\
        \midrule
        1  & 93.34 & 99.63 \\
        5  & 92.56 & 99.70 \\
        \textbf{10}  & \textbf{93.60} & \textbf{99.73} \\
        20 &  92.73 & 99.68 \\
    \bottomrule
    \end{tabular}
	    \end{minipage}
	\hfill
	\begin{minipage}{0.45\linewidth}
		\centering
		\setlength{\abovecaptionskip}{0.28cm}
		\includegraphics[width=2.6cm]{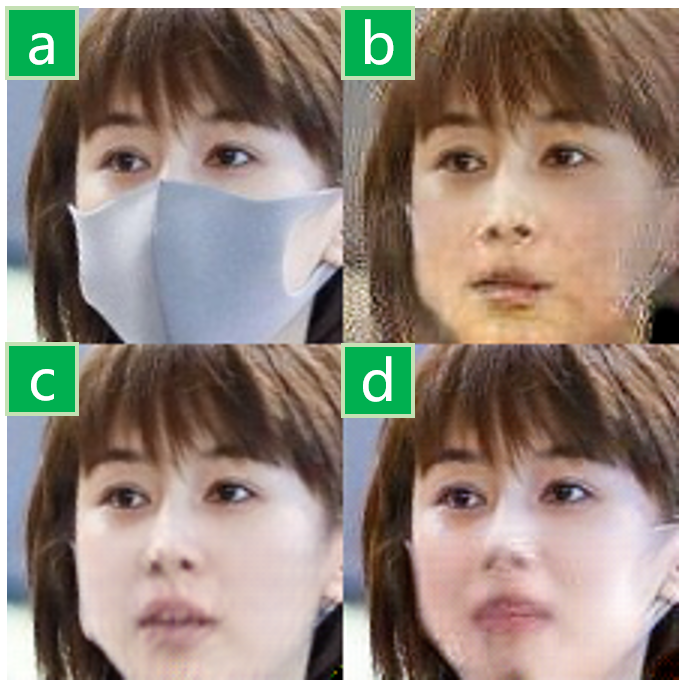}
	\end{minipage}
\caption{The accuracy(\%) of different hyper-parameter $\gamma$ in Eqn.\ref{l_total}. The right part gives the generated samples with different $\gamma$. Image a is the original image, and b, c, d are the generated images with $\gamma=$ 1, 10 and 20.}
\label{tab+fig1}
\end{figure}

For the hyper-parameters, we compare the performance of different values of $\gamma$ and $\eta$ in Eqn.\ref{l_total}. In the jointly training of $E$, $D$ and $D_{adv}$ in MEER-stage2 , the ratio of reconstruction loss and GAN adversarial loss determines 
whether the constraints of the generative model are more explicit or more realistic.
In this experiment, we fix the factor of adversarial loss $\mathcal{L}_{adv}$ and change the hyper-parameter $\gamma$ of reconstruction loss. 
Illustrated in the left table of Figure~\ref{tab+fig1}, 
when a balanced $\gamma$ is adopted ($\gamma=10$), the accuracy on both RMFD and LFW gets the best. 
As shown in the right of Figure~\ref{tab+fig1}, when the value of $\gamma$ is small (i.e. 1), the generated network will be difficult to train, and it produces image with artifacts around hair and mouth (b). When the value of $\gamma$ is large (i.e. 20),
the generated image becomes blurry (d). And generated image (c) with balanced $\gamma=10$ gets the best qualitative result.



\begin{figure}[htbp]
    \centering
	\begin{minipage}{0.45\linewidth}
		\centering
        \begin{tabular}{cccc}
        \toprule
        \textbf{$\eta$} & \textbf{RMFD} & \textbf{LFW} \\
        \midrule
        0 & 92.54 & 99.70 \\
        0.05 & 93.16 & 99.73 \\
        \textbf{0.1} & \textbf{93.60} & \textbf{99.73} \\
        0.5 & 92.93 & 99.71 \\
        1   & 93.12 & 99.71 \\
        \bottomrule
        \end{tabular}
        \label{fig:NormalSafe}
	    \end{minipage}
	\hfill
	\begin{minipage}{0.45\linewidth}
		\centering
		\setlength{\abovecaptionskip}{0.28cm}
		\includegraphics[width=\linewidth]{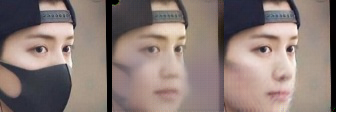}
	\end{minipage}
\caption{The accuracy(\%) of different hyper-parameter $\eta$ in Eqn.\ref{l_total}. Note that $\eta=0$ means no id-preserving loss used in stage 2 training, whose accuracy values on both RMFD and LFW are lower than those of $\eta=0.1$. The right part is a generated image comparison with different $\eta$. From the left to the right, they are original masked image, generated image with and without  id-preserving loss ($\eta=$ 0 and 0.1).}
\label{tab+fig2}
\end{figure}

At last, a similar experiment is performed on different factors of id-preserving loss. As shown in the left table of Figure~\ref{tab+fig2}, the hyper-parameter $\eta$ of the  id-preserving loss is optimal at 0.1. 
$\eta=0$ means that identity preserving is not used, and the corresponding performance falls by 1.06\% on RMFD. 
As shown in right of Figure~\ref{tab+fig2}, 
from left to right, it shows the original masked image, the generated image with and without
id-preserving loss ($\eta=$ 0 and 0.1).
Identity of the middle face without id-preserving loss can not be guaranteed, where it is less similar with the left original face than the rightmost face. 

\section{Conclusion}
In this work, we proposed a multi-task learning framework, termed MEER, to achieve MFR and face mask removal.
We proposed a novel decoupling strategy to achieve purify face identity feature and restore masked facial part based on joint training strategy. 
And we build an id-preserving loss to further keep ID consistence between original real unmasked face and its corresponding generative mask removal face.
Extensive experiments on masked face datasets demonstrate superiority of the proposed method.

{\small
\bibliographystyle{ieee_fullname}
\bibliography{egbib}
}

\end{document}